\documentclass[letterpaper, 10 pt, conference]{ieeeconf} 
\IEEEoverridecommandlockouts
\usepackage{graphics} 
\usepackage{epsfig} 
\usepackage{mathptmx} 
\usepackage{times} 
\usepackage{amsmath} 
\usepackage{amssymb}  
\usepackage[at]{easylist}
\usepackage{setspace}
\usepackage{xcolor}
\usepackage[binary-units=true]{siunitx}
\usepackage{mathtools}
\usepackage{float}
\usepackage[ruled,vlined,linesnumbered,noresetcount]{algorithm2e}
\usepackage[pagebackref=true,breaklinks=true,letterpaper=true,colorlinks,bookmarks=false]{hyperref}
\usepackage{multirow}

\usepackage{etoolbox}
\makeatletter
\patchcmd{\@makecaption}
  {\scshape}
  {}
  {}
  {}
\makeatother

\title{\LARGE \bf Gel\textit{Slim} 3.0: High-Resolution Measurement of Shape, Force and Slip\\
in a Compact Tactile-Sensing Finger}

\author{
\authorblockN{Ian H. Taylor$^1$, Siyuan Dong$^2$, and Alberto Rodriguez$^1$}
\authorblockA{{\tt\small <ihtaylor, sydong, albertor>@mit.edu}}
\authorblockN{ \href{https://github.com/mcubelab/gelslim}{https://github.com/mcubelab/gelslim}}
\thanks{\hspace{-0.3cm}$^1$Dept. of Mechanical Engineering, Massachusetts Institute of Technology \newline $^2$Dept. of Electrical Engineering and Computer Science, Massachusetts Institute of Technology \newline
Project repository: \href{https://github.com/mcubelab/gelslim}{https://github.com/mcubelab/gelslim} \newline
Project video: \href{https://www.youtube.com/watch?v=Y10XN9byO0g}{https://www.youtube.com/watch?v=Y10XN9byO0g}}
}

\newcommand{\secref}[1]{Section~\ref{#1}}

\begin{document}
\maketitle

\begin{abstract}
This work presents a new version of the tactile-sensing finger GelSlim 3.0, which integrates the ability to sense high-resolution shape, force, and slip in a compact form factor for use with small parallel jaw grippers in cluttered bin-picking scenarios. The novel design incorporates the capability to use real-time analytic methods to measure shape, estimate the contact 3D force distribution, and detect incipient slip. To achieve a compact integration, we optimize the optical path from illumination source to camera and other geometric variables in a optical simulation environment. In particular, we optimize the illumination sources and a light shaping lens around the constraints imposed by the photometric stereo algorithm used for depth reconstruction. The optimized optical configuration is integrated into a finger design composed of robust and easily replaceable snap-to-fit fingetip module that allow for ease of manufacture, assembly, use, and repair. To stimulate future research in tactile-sensing and provide the robotics community access to reliable and easily-reproducible tactile finger with a diversity of sensing modalities, we open-source the design and software at \href{https://github.com/mcubelab/gelslim}{https://github.com/mcubelab/gelslim}.
\end{abstract}

\section{Introduction}
\label{sec:intro}
Touch is an essential sensing modality for interfacing with the world. It provides a direct and powerful mechanism for reacting to the environment, present in almost the entirety of the animal kingdom but missing from most of the robotic one.
%
%
We are particularly motivated by the potential of tactile feedback to sense and control the complex contact interactions that occur while grasping, using, and placing objects within an environment. 

This paper presents GelSlim 3.0, a new version of the tactile-sensing finger GelSlim~\cite{donlon2018gelslim}, designed for grasping in cluttered environments, and integrating the real-time measurement with high spatial resolution of shape, force, and slip, in a compact form factor that is easy to fabricate, utilize and maintain. We present four main contributions. 

%
\begin{figure}[h]
\label{fig:figure1}
    \centering
    \includegraphics[width=\columnwidth]{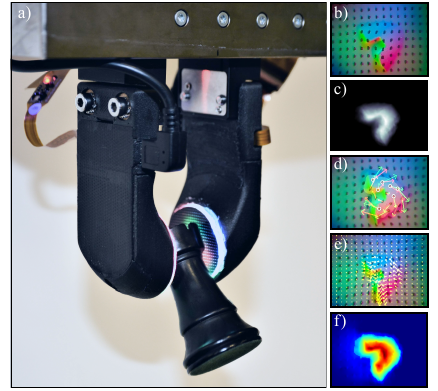}
    \vspace{-.5 cm}
    \caption{(a) GelSlim 3.0 sensors grasping a chess rook and the respective tactile imprint and measurement results. (b) High-Resolution Tactile Image (c) Depth Reconstruction (d) Incipient Slip (e) Tangential Force Field (f) Normal Force Field.  }
    \label{fig:introduction}
    \vspace{-.5 cm}
\end{figure}

\begin{itemize}
\item\textbf{Design:} 
%
GelSlim 3.0 is product of an optimization of the optical path, from illumination source to camera, around the constraints imposed by the photometric stereo algorithm used for depth reconstruction in vision-based tactile sensors~\cite{improved_gelsight}. In particular, we optimize the illumination sources and the geometry of a light shaping lens to enable a compact form factor integration while retaining the ability to reconstruct depth. We evaluate the novel tactile-sensing finger compared to current tactile sensor designs and discuss our implementation's trade-offs and limitations.

\item\textbf{Manufacturing:} We employ accessible manufacturing methods to fabricate the new tactile-sensing finger, which is assembled out of a small number of components with a simple and modular snap-to-fit design.

\item\textbf{Functionality:} The tactile sensor integrates the capability to use real-time analytical methods to measure 3D geometry reconstruction, to detect incipient slip, and to estimate the spatial distribution of 3D contact forces.

\item\textbf{Open-Source:} We open-source the design 
to help address some of the many bottlenecks in tactile-sensing in robotics, which include manufacturing, distribution, and use \cite{digit}, and to encourage the proliferation of tactile-sensing technology within the robotics community. 
\end{itemize}

This paper is structured as follows: We review related works in tactile-sensing in \secref{sec:lit_review} and discuss the design goals in \secref{sec:goals}. We describe the optimization and integration of the optical, electrical, and mechanical design components in \secref{sec:design} and demonstrate the sensing capabilities of resultant tactile-sensing finger in \secref{sec:algorithms}. We compare the novel design to other current tactile-sensing designs and discuss its limitations in \secref{sec:discussion}. Finally, we briefly discuss the open-source design repository in \secref{sec:source} and summarize the contributions of the paper in \secref{sec:conclusion}. 
\section{Related work}
\label{sec:lit_review}
We review relevant work regarding the design of vision-based tactile sensors that utilize analytical and learning-based methods of tactile data interpretation.

\begin{figure}[H]
    \centering
    \includegraphics[width=\columnwidth]{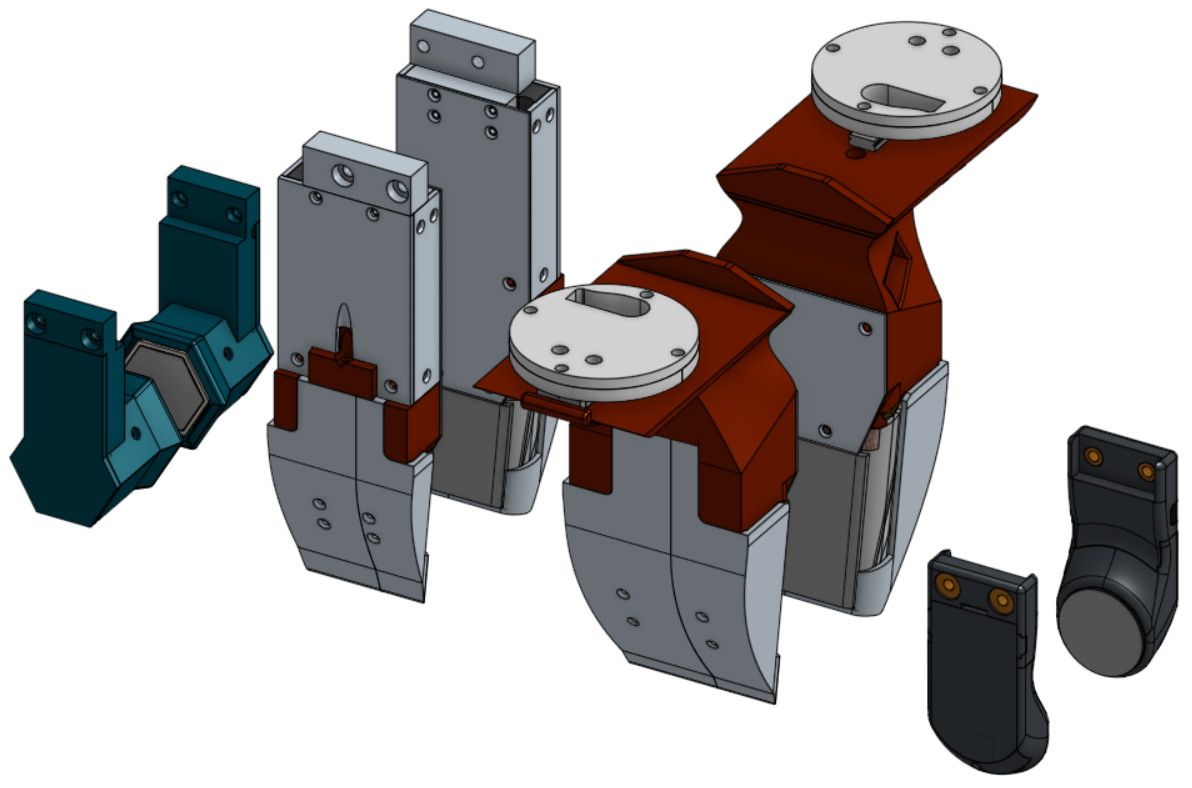}
    \vspace{-.5 cm}
    \caption{Models of previous and current integrated tactile sensors (\textit{from left to right}): GelSight \cite{improved_gelsight}, GelSlim 2.0 \cite{dense_force}, GelSlim MPalm \cite{dexterity}, and GelSlim 3.0. }
    \label{fig:versions}
\end{figure}


\subsection{Vision-Based tactile-sensing}

Vision-based retrographic tactile sensors have become popular within the robotics literature due to their ability to capture contact state feedback with high spatial resolution while maintaining a compact instrumentation \cite{donlon2018gelslim}. These approaches are also synergistic with deep learning techniques, which have been demonstrated to provide robust and descriptive interpretations of high-dimensional data. The continued development and improvement of these sensors have yielded several designs that offer compactly-integrated and rugged form factors \cite{donlon2018gelslim,digit}, where previously they had been more bulky and/or fragile \cite{bio_encoding, localization, improved_gelsight}.


\subsection{GelSight Sensors}
GelSight is a vision-based tactile sensor that makes measurements by using a camera to capture high-resolution images of the surface deformation of an elastomer with opaque skin. The elastomer is illuminated from different directions by color LEDs; the resultant colored shading is used to make direct measurements of the position, shape, and 3D geometry of objects contacting the surface. A detailed review of different versions of GelSight sensors can be found in \cite{improved_gelsight}. Donlon et al. \cite{donlon2018gelslim,dense_force} improved upon Dong et al.’s \cite{improved_gelsight} fingertip version of GelSight by altering the optical path to use a waveguide to route light through the fingertip and a mirror to view the sensor surface from a distance, allowing for a more compact wedge-shaped fingertip design, suitable for cluttered picking scenarios. Wilson et al. \cite{fullyactuatedgelsight} demonstrated a fully actuated two finger-gripper with multiple GelSight tactile sensors, which were used to gather the surface topology of an object across multiple views simultaneously and track the shear and and tensile stresses. Sensors similar to GelSight and GelSlim that utilize rich 2D and 3D data have been successfully applied in robotic manipulation. Li et al. \cite{smallparts} utilized GelSight fingertip in tandem with a feature-based matching technique to localize and perform an insertion task with small parts held in a robot hand. Tian et al. \cite{Manipulationbyfeel} Proposed and demonstrated an unsupervised learning framework, deep tactile MPC,  to learn a predictive model from raw, tactile sensor inputs for performing servoing tasks with a GelSight-style tactile sensor. Hogan et al. \cite{dexterity} explored the use of tactile feedback measured by a palm-shaped GelSlim sensor to develop robust primitives with the ability to handle external perturbations and account for object pose uncertainty. Bauza et al. \cite{bauza_localization} utilized GelSlim sensors to reconstruct the shape of objects from tactile imprints and accurately localize previously reconstructed objects. She et al. \cite{cable_manipulation} demonstrated the ability to manipulate flexible cables by pairing grip and pose controllers that utilized depth and shape feedback from a GelSight sensor to regulate cable sliding dynamics.


\subsection{Other Vision-Based Tactile Sensors} 
There is a diversity of vision-based tactile-sensing designs that rely more heavily on machine learning-based data interpretation methods. For example, She et al. \cite{exoskeleton} demonstrated an exoskeleton covered soft continuum finger that utilized a convolutional neural network to predict the finger's position. Padmanabha et al. \cite{omnitact} used a thumb-shaped tactile sensor, OmniTact, paired with a ResNet-based neural network to perform tactile state estimation for a representative insertion task.
While the number of vision-based tactile-sensing designs continues to increase, Lambeta et al. \cite{digit} notes the dearth of sensors that simultaneously fulfill the requirements of being 1) high resolution, 2) highly sensitive, 3) reliable, 4) easy to use, 5) compact and 6) inexpensive. To resolve this bottleneck impeding the widespread adoption of touch sensing in robotic manipulation, they released an open-source tactile sensor Digit that fulfills these criteria.

\section{Design Goals}
\label{sec:goals}

In previous versions of GelSight and GelSlim, a clear gel with an opaque outer membrane is illuminated by a light source and captured by a camera. The position of each of these elements depends on the specific requirements of the sensor \cite{donlon2018gelslim}. A primary influence on the configuration is whether the tactile sensor incorporates model-based photometric stereo techniques to measure 3D geometry. The decision to incorporate these techniques constrains the design space as it requires that at least three colors of light be directed across the gel from different directions \cite{retrographic}. These requirements also enforce geometric constraints that affect feasible camera placement and illumination paths. For example, the need for multiple evenly distributed illumination channels necessitates designs that position luminaries in a radially symmetric configuration. Limitations to the focal length of available cameras constrain the maximum field of view possible, and thus, sensing area for a given sensor thickness. A comprehensive review of the tactile design space constraints can be found in \cite{donlon2018gelslim}. These constraints often prove challenging and counterproductive to slim robotic finger integrations \cite{donlon2018gelslim}. While previous implementations of GelSlim have eschewed these techniques in favor of relaxing the aforementioned design space constraints \cite{donlon2018gelslim,dense_force}, these techniques have and continue to prove useful for complex dexterous manipulation scenarios \cite{cable_manipulation}.

Previous designs of GelSight and GelSlim, while cost-effective, have been difficult to manufacture and assemble en-masse or by inexperienced users due to their reliance on complex by-hand fabrication techniques that are incompatible with retail fabrication services (e.g., acrylic waveguide bending, custom hand-soldered luminaries)\cite{digit,donlon2018gelslim,li_sensor,improved_gelsight}. 

To address these issues while maintaining the strengths of previous GelSight and GelSlim designs, we present the following design goals for the proposed tactile finger:
\begin{itemize}
    \item \textbf{Compactness} of the finger allows access to and precise placement of objects in clutter by squeezing between objects or separating them from the environment.
    \item \textbf{Analytic Measurement Methods} utilized in previous high-resolution camera-based tactile sensors provide multiple types of rich contact state feedback (raw high-resolution image, shape, force, slip) at real-time speeds without requiring data-expensive calibration or learning procedures. These provide a strong basis for novel methods of dexterous manipulation and control. 
    \item \textbf{Design for Assembly (DFA)} enforces robust and modular design practices that make the fabrication and utilization of the proposed tactile finger accessible to a broader range of the robotics research community.
\end{itemize}

The following sections give a detailed description of the process to use these design goals to select and optimize the optical (\secref{subsec:optical}), electrical (\secref{subsec:electrical}), and mechanical (\secref{subsec:mechanical}) design parameters.
Finally, we describe the fabrication and assembly of the tactile finger (\secref{subsec:fabrication}).

\section{Design of GelSlim 3.0}
\label{sec:design}

\begin{figure*}[t]
    \centering
    \includegraphics[width=\textwidth]{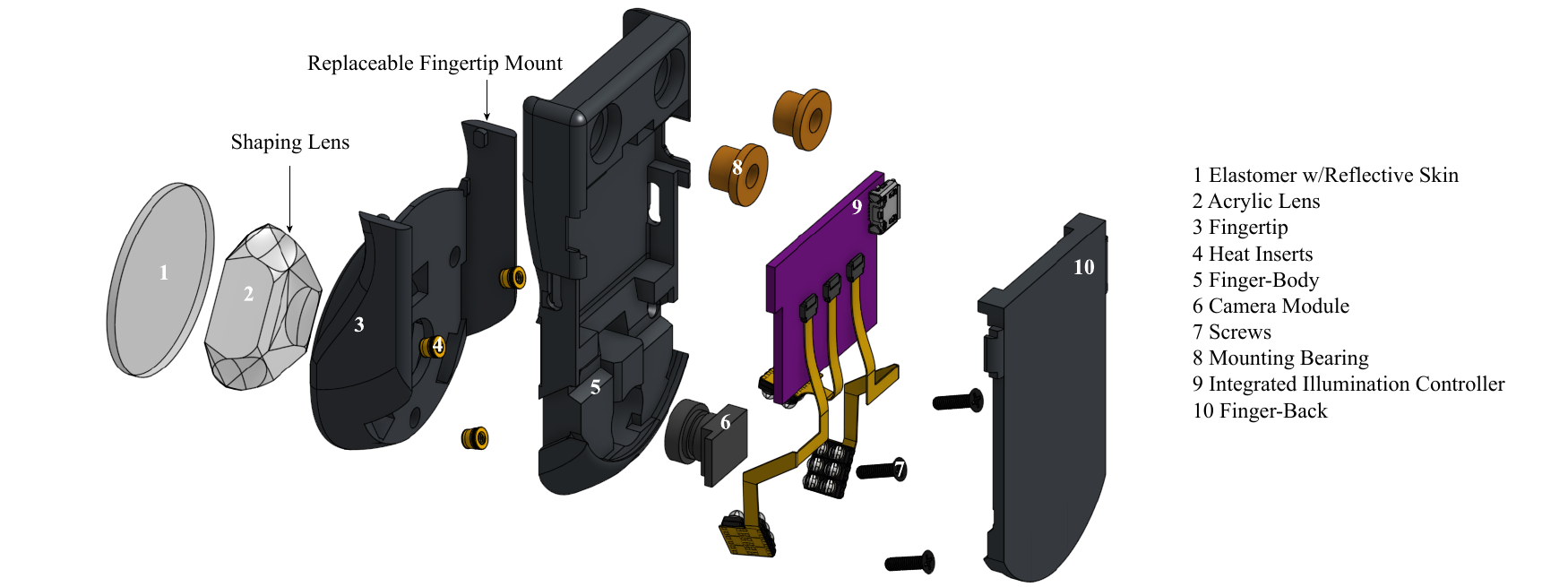}
    \vspace{-0.5 cm}
    \caption{ Exploded diagram of the GelSlim 3.0 tactile finger and its 10 components}
    \label{fig:exploded}
    \vspace{-.5 cm}
\end{figure*}

\subsection{Optical Design}
\label{subsec:optical}
The use of photometric stereo techniques is a primary influence on the constraints of a tactile sensor's optical design space. Therefore, we briefly review the photometric stereo algorithm and its assumptions (\ref{subsubsec:photometric}). We evaluate these restrictions paired with the proposed design goals to set the initial conditions for the design space (\ref{subsubsec:initial}). We then generate a raytracing simulation of an initial optical path using illumination design software, identify design optimization parameters (\ref{subsubsec:simulation}) and generate an optimized optical configuration using the software's built-in optimizer (\ref{subsubsec:optimization}). \vspace{.2 cm}

\subsubsection{Photometric Stereo}
\label{subsubsec:photometric}
Our implementation of photometric stereo relies on the same techniques first described by Johnson et al. \cite{retrographic} and used in the original GelSight sensor designs as a way to analytically reconstruct the deformation of an elastomer in 3D. The technique assumes the following three conditions: 

\begin{enumerate}
 \item The image seen by the vision system is an orthographic projection of the sensor surface, and thus every point in the image corresponds to a point on the sensor surface. Thus, the gradients of the surface deformation can be defined as the partial derivatives of the points seen in the image.
 
\item The shading at any point in the image is a function of the surface normal of the sensing surface; this effectively assumes no cast shadows or interreflections in the image.

\item The albedo (the proportion of incident light reflected by the contact surface) is constant across the sensor's surface. 
\end{enumerate}

To reconstruct the surface normals we determine the mapping between the intensity at each pixel and the reflectance function. Theoretically, many sets of gradients will correspond to a set of intensity values; therefore, the reflectance function is not ``trivially invertible." For this reason, the photometric stereo technique uses images captured under different illumination conditions (from three differently colored channels). The three measurements per pixel overconstrain the problem and allow us to estimate the two gradient values. We measure the intensities using an object of a known shape, e.g., a sphere with a known radius. From the data, we generate a lookup table; this serves as the method of inversion. Finally, we use a fast Poisson solver to integrate the surface gradient and reconstruct the depth.

While this method does not assume point light sources \cite{retrographic}, we incorporate illumination homogeneity as a second-order assumption. Inhomogeneous behavior, e.g., areas with intensity variance and missing illumination, will reduce the depth reconstruction accuracy.  The former will disproportionately skew the gradient functions, and the latter will reduce the number of measurements per pixel, making the reflectance-intensity mapping not trivially invertible and will generate incorrect depth values. \vspace{.2 cm}

\subsubsection{Initial Conditions}
To satisfy the photometric stereo assumptions, we select initial design parameters based on previous designs of GelSight and GelSlim as well as our design goals. We discuss the effects our selections have on the design space.
\label{subsubsec:initial}

\begin{itemize}
\item\textbf{Camera:} We choose a Raspberry Pi 160{\textdegree} variable focus wide-angle camera module (ODSEVEN) as it is an inexpensive and compact camera that provides a large field of view and a close minimum focus distance. We cannot completely satisfy the orthographic assumption as the camera's output is warped due to the spherical distortion induced by the wide-angle lens; thus, while all points in the image correspond to a point on the sensing surface, they are not orthographic projections of one another. We address this discrepancy in \secref{subsec:process} by performing a pre-processing step that removes the wide-angle distortion.

\item\textbf{Shaping Lens:} We select a hexagonal prism with an angled loft as the basis for the initial lens geometry. The hexagonal shape naturally satisfies the needs of the photometric stereo algorithm used with three illumination channels by providing radially symmetric illumination, which induces the required constraints on the reflectance function \cite{retrographic}.  We scale the initial lens geometry to maximize the camera modules’ field of view and, thereby, the total sensing area while constraining the maximum finger thickness to be 20 mm. We choose this constraint based on our observations of compact grippers commonly used in industrial pick-and-place applications.  

\item\textbf{Gel Shape:} The light distribution is affected by the sensing surfaces’ curvature. Tactile-sensing applications often use parabolic and spherically curved as well as flat planar sensing surfaces. In some instances, curved surfaces have been used as a replacement for complex shaping lens geometry \cite{improved_gelsight}. We choose to use a planar surface to allow the fingers to more readily grasp objects lying flush against flat surfaces, for example, on tables or at the bottom of picking bins \cite{pick_and_place}. 

\item\textbf{Gel Material:} We choose to use an elastomer skin with Lambertian reflectance that provides a uniform and diffuse albedo. This enables measurements of the surface normal which are more accurate than those where a semi-specular albedo is used \cite{improved_gelsight}.

\item\textbf{Illumination}
Using a Lambertian reflectance skin limits the design space as the intensity of light reflected by the diffuse albedo is too low to effectively use total internal reflection (TIR) techniques to route and distribute the light across the surface; a technique used in previous versions of GelSlim and GelSight with semi-specular skin \cite{donlon2018gelslim,li_sensor}. As a result, the luminaries must directly illuminate the elastomer's surface, and the optical design must incorporate alternative methods to illuminate the sensor's surface homogeneously. We aim to accomplish this by optimizing the light distribution with an illumination shaping lens.\vspace{.2 cm}

\end{itemize}
\subsubsection{Simulation and Parameter Selection}
\label{subsubsec:simulation}
We simulate the optical system configured by the initial conditions using \href{https://www.synopsys.com/optical-solutions/lighttools.html}{Synopsys LightTools Illumination Design Software} linked to a Solidworks CAD model of the shaping lens; this allows us to use Solidworks geometric feature parameters directly as optimization variables. To accurately simulate the optical system, we consider the following:


\begin{itemize}

\item\textbf{Material Parameters:} To model the ray interactions between different materials, we set the refractive indices and reflectance of each optical component using built-in material profiles.  
\item\textbf{Data Collection:} We place a  forward simulation receiver on the modeled sensing surface to collect intensity and color distribution data. 
\item\textbf{Source Models:} To simulate the selected photodiodes, we use source models provided by the luminary manufacturer, which contain their package CAD, source distribution, and spectral region data.
\item\textbf{Emittance:} To model an intensity-matched ideal photometric stereo system, we match radiometric power for all sources.

\end{itemize}

\begin{figure}[h]
    \centering
    \includegraphics[width=\columnwidth]{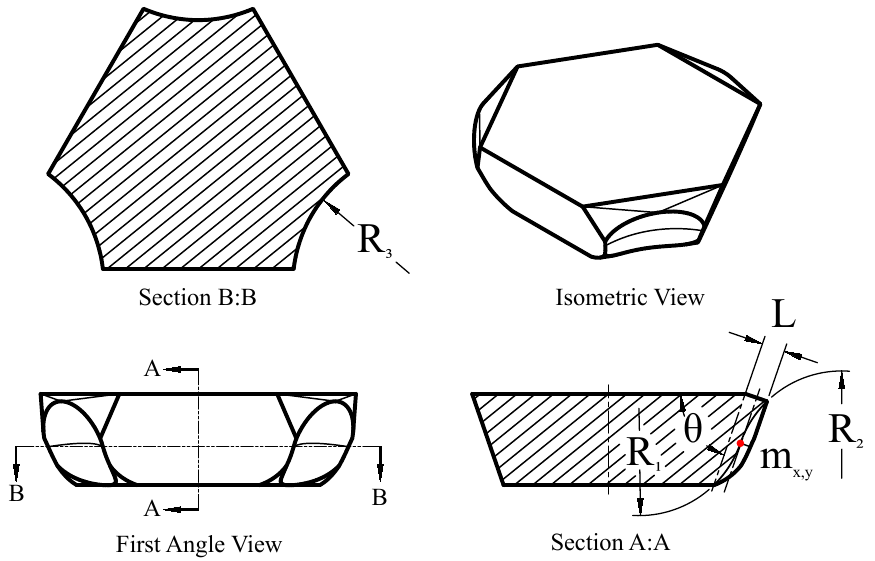}
    \vspace{-.5 cm}
    \caption{Diagram of the hybrid swept spline lens with optimization parameters labeled across multiple views. The point $m_{x,y}$ is marked with a red dot.}
    \label{fig:lens}
    \vspace{-.5 cm}
\end{figure}

\begin{figure}[h!]
    \centering
    \includegraphics[width=\columnwidth]{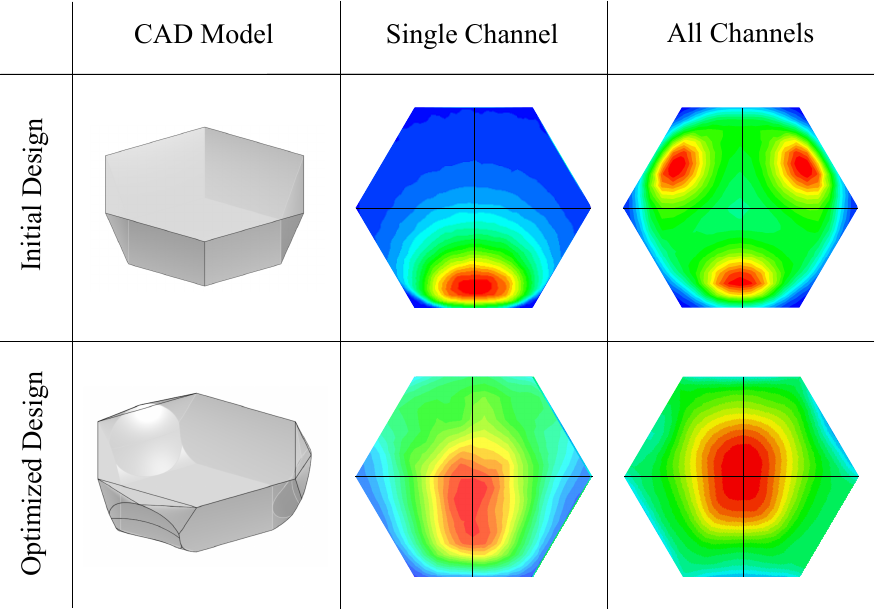}
    \vspace{-.5 cm}
    \caption{Heat-maps of the simulated radiant flux across the surface of the sensor before and after optimization plotted using a $25 x 25$ bin receiver mesh.}
    \label{fig:heatmap}
    \vspace{-.25 cm}
\end{figure}

We perform proof of concept testing to constrain the design space and select the lens shape and LED package type. We find that a hybrid swept spline lens distributes illumination intensity more homogeneously than a lens with flat sides \ref{fig:heatmap} while inducing fewer interreflective visual artifacts than planar or spherical convex and concave lenses. Additionally, we find that a 60\textdegree~viewing angle PLCC-4 Package OSRAM TOPLED Black T66 series LED provides an illumination profile that is diffuse yet easily shaped. We initialize the optimization in Eq.(\ref{eq:optimization}) according to the features of the design space seen in Fig.~\ref{fig:lens}:

\begin{align}
    \label{eq:optimization}
    \raggedright
    \noindent
    &Given~~\mathbf{\overline{A}=[x,y,z,\alpha],~\overline{B}=[L,\theta,R_{1},R_{2},R_{3}]}~~~~~~~~~~~\\
    \nonumber &minimize~~f(\overline{A},\overline{B})
\end{align}
\vspace{-.675 cm}
\begin{table}[htb]
\begin{tabular}{ll}
$subject~to:$ \kern .166em $\mathbf{0 \leq z\leq t_{1}}$ ~~~$and$  & $\mathbf{\sigma(x,y)=(0,0)}$                                               \\
~~~~~~~~~~~~~~~$\mathbf{0 \leq L \leq t_{2}}$ &  $\mathbf{\mu_{CIE}(x,y)=(.33 \overline{3},.33 \overline{3})}$                    \\
~~~~~~~~~~~~~~~$\mathbf{0\leq \alpha \leq 90}$ &   $\mathbf{G(x,y)=(0,0)}$                                                          \\
~~~~~~~~~~~~~~~$\mathbf{0\leq \theta \leq 90}$ &  $\mathbf{x=m_{x},~ y\geq m_{y}}$                                             \\
                                &  $\mathbf{R_{1}\geq 0}, ~\mathbf{R_{2}\geq 0}, ~ \mathbf{{R_{3}\geq 0}}$
\end{tabular}
\end{table}
\vspace{-.25 cm}

Where $\overline{A}$ and $\overline{B}$ represent the LED position and shaping lens parameters, respectively, where $x,y,z$ are the coordinate positions within cartesian space and $\alpha$ is the angle of the LED relative to the top face of the lens. $t_{1}$ is the total thickness of the lens, and $t_{2}$ is the maximum thickness of the shaping feature. The maximum thickness is constrained to 2 mm to ensure the shaping feature does not extend too far from the sensing surface and increase the finger's total thickness. $m_{x,y}$ is coordinate origin within the shaping feature plane. 

In abstract, the cost function $f(\overline{A},\overline{B})$ represents the design space modeled by the initial optical configurations within the simulator engine. The CIE mean ($\mu_{CIE}$) represents the average chromaticity coordinate value for all the bins (i.e., test coordinates)  in a receiver mesh,  where the X coordinate represents a mix of the three CIE RGB curves and Y represents the luminance value. The goal value designates the chromaticity coordinates of the reference white point (the color coordinates that defined the color "white" and thus, the point at which red, green, and blue are mixed homogeneously). Satisfying this goal ensures that the \textit{total} illumination distribution approximates the white point. The illuminance mesh standard deviation ($\sigma$) goal value incentivizes the intensity to remain constant across the sensing surface. Thus, the optical configuration will maintain a homogeneous and evenly mixed distribution. The illumination centroid ($G(x,y)$) is the position of the arithmetic mean of the intensity distribution across the local receiver bins. The goal value further enforces illumination homogeneity by addressing assumption (2). The goal value incentivizes positioning the centroid at the center of the lens ($G(x,y)=(0,0)$). This directly addresses assumption (2) as it reduces interreflections, which most often occur when high-intensity illumination refracts and reflects while directed at the acute features (ex: sharp corners) often present near the boundary of a lens. Due to the optical configuration's radial symmetry, we can utilize a single parameter set to represent the shaping geometries used for each of the three individual color channels. The full optical configuration is used during the optimization of the proposed minimization problem. \vspace{.2 cm}


\subsubsection{Optimization Results}
\label{subsubsec:optimization}
The optimizer evaluates as a two-step process where Solidworks configures the lens features based on the optimization variables and initial conditions, and LightTools performs raytracing to simulate the optical system and evaluate the cost function. We plot and display a comparison of intensity distributions of the initial and final geometries in Fig.~\ref{fig:heatmap}. The optimized shaping features effectively distribute the illumination across the surface of the sensor. We observe that the full channel distribution of the optimized lens is not entirely radially symmetric; this is likely due to interreflective artifacts generated within the compact form factor lens. Additionally, while the distribution is improved, it is not ideally homogeneous as there are still portions of the sensing surface near the edges where there are large deviations in the illumination. We find that at the origin coordinate, the optimized design sufficiently satisfies the aforementioned first and second-order assumptions and demonstrates the desired ability to use the photometric stereo algorithm to generate 3D reconstructions with low noise as seen in Fig.~\ref{fig:figure1} and Fig.~\ref{fig:reconstruction}. Reconstructions at the peripheral edges of the sensor are more prone to noisy behavior. This is in comparison with the initial design, which did not produce usable photometric data. We further discuss the finger's resultant capabilities and compare the performance to previous tactile sensor designs in \secref{sec:discussion}.

\begin{figure}[h!]
    \centering
    \vspace{-.25 cm}
    \includegraphics[width=\columnwidth]{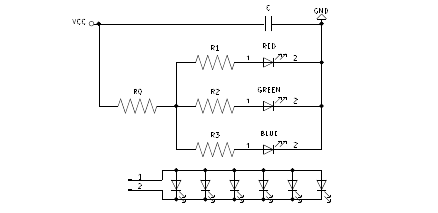}
    \vspace{-.75 cm}
    \caption{Circuit diagrams of the Driver PCB (Top) each color channel has its own LED array (Bottom). }
    \label{fig:circuit}
    \vspace{-.5 cm}
\end{figure}

\subsection{Electrical Design}
\label{subsec:electrical}
  The electrical assembly supplies power to and secures the LED luminaries. The circuit diagram is shown in \ref{fig:circuit}. The resistance values R1, R2, and R3, normalize the three color channel's relative illumination while R0 modulates the total intensity. We use the camera module's spectral transmission profile and manufacturer specifications of the current vs. luminous intensity relationship for each LED to select and tune the channel resistances. The LEDs for each channel are wired as a 3x2 array in parallel to homogenize the illumination and minimize sequential intensity loss across LEDs in a channel.


\subsection{Mechanical Design}
\label{subsec:mechanical}
The mechanical assembly consists of the fingertip, finger-body, and finger-back. These three components are designed to encapsulate the optical path and provide the most compact form factor possible. The fingertip's distal face is curved, which allows the finger to more readily slip between objects when grasping in clutter. The mounting geometry is designed for use with a Weiss WSG-32 gripper but can be easily adjusted to accommodate various grippers. The fingertip and finger-body are joined with an easily separable sliding rail system in addition to several screws. The respective component cavities of the two bodies maintain press-fit tolerances. In combination, these features allow for easy assembly and maintenance of the sensor, as any component can be accessed and easily removed for a replacement if necessary. The modular snap-to-fit mechanical design allows the fingertip to be easily swapped out, if necessary, to switch between task-specific elastomers.


\subsection{Fabrication and Assembly}
\label{subsec:fabrication}
The finger assembly is composed of ten components as seen in Fig.~\ref{fig:exploded}, all of which, except the elastomer, are either 3D printed (Fingertip, finger-body, finger-back), retail made-to-order (Driver PCB, Illumination harness, and acrylic lens) or off the shelf components (camera module, mountain bearings, and M2 Heat Inserts and screws). Once the heat inserts are embedded in the fingertip using a soldering iron, the finger is quickly assembled because of the modular "snap-to-fit" design.

The elastomer is cast using XP565 (Silicones Inc.) platinum-based silicone mixed in a 1:10 ratio of activator to base. After degassing in a vacuum chamber, it is poured into an acrylic mold. Once cured, the reflective layer was painted using a gray silicone ink (Print-On Silicone Ink, Raw Material Suppliers) using a 10:1 ratio of silicone ink to its catalyst. A silicone solvent (NOVOCS Gloss, Reynolds Advanced Materials) is added to dilute the silicone ink in a 1:10:30 ratio of ink catalyst to ink to solvent, and an airbrush is used to spray the material on top of the elastomer. After the reflective layer has cured, tracking markers are added to the elastomer by laser cutting holes in a grid pattern on the elastomer's painted surface and adding black silicone ink. The marker layer is sealed by spraying an additional layer of black silicone ink in a process similar to that of the reflective layer. Finally, the elastomer is cut to shape using a template and then bonded to the acrylic lens and finger-body using silicone sealant (Gorilla Silicone Clear Sealant); the completed sensor is then left to cure for 24 hours.


\section{Integrated Algorithms for tactile-sensing}
\label{sec:algorithms}
We briefly discuss the image processing pipeline and demonstrate the measurement methods enabled by the novel GelSlim sensor design.
\subsection{Pre-Processing}
\label{subsec:process}
To sense physically meaningful contact with the tactile finger, one pre-processing step is needed to correct the optical distortion introduced by the acrylic shaping lens (Part 2 in Fig.~\ref{fig:exploded}) and the wide-angle camera module. This wide-angle causes distortions in the raw, tactile image output which can be clearly visualized by the ``curved" marker array in the second image of Fig~\ref{fig:processing} below. To correct the distortion, we first generate the binary mask from the raw, tactile image to separate the marker array from the background and detect each marker's position. Using the marker array's known distribution, we calculate the correspondence between each marker in the distorted image and its true position. We interpolate this correspondence across all of the pixels in the raw image to find the transformation matrix that will reverse the distortion. Finally, we apply the transformation matrix to correct the raw image. 

\begin{figure}[H]
    \centering
    \includegraphics[width=\columnwidth]{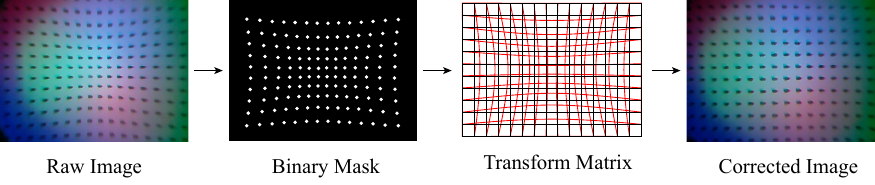}
    \vspace{-0.5 cm}
    \caption{Diagram of the processing pipeline used to correct spherical distortion. }
    \label{fig:processing}
    \vspace{-.25 cm}
\end{figure}

\subsection{Shape and Geometry}
\label{subsec:shape_geometry}

\begin{figure}[h!]
    \centering
    \includegraphics[width=\columnwidth]{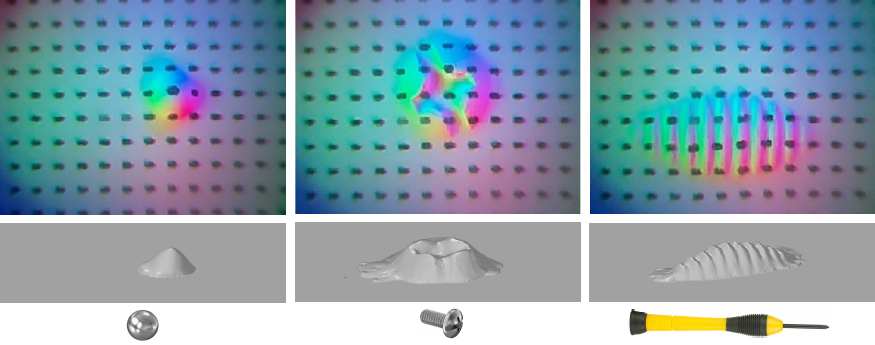}
    \vspace{-0.5 cm}
    \caption{Raw tactile imprints, prior to pre-processing, and their respective 3D reconstructions \textit{(from left to right)}: ball bearing, screw head, screwdriver handle.}
    \label{fig:reconstruction}
    \vspace{-.5 cm}
\end{figure}

We describe the photometric stereo depth reconstruction algorithm and its assumptions previously in \secref{subsubsec:photometric}. Fig.~\ref{fig:reconstruction} demonstrates these capabilities and shows the measured RGB image and reconstructed depth map of some daily objects touched. The local 3D geometry of the object measured by GelSight/GelSlim sensor has been demonstrated to be useful for object pose estimation~\cite{cable_manipulation,bauza_localization} and force distribution estimation~\cite{dense_force}.%

\subsection{Shear and Incipient Slip}
\label{subsec:shear_slip}

An approximated shear force~\cite{cable_manipulation} can be estimated by tracking the motion of the marker array. By analyzing the coherency of the marker motion field, we can also predict slip. The method proposed by Dong et al.~\cite{gelsight_incipient} utilized with the previous GelSlim design is compatible with the GelSlim 3.0 finger with minor changes. 

The core idea is to distinguish whether the sensor surface's motion during contact is consistent with a rigid body motion or not. When the object and sensing surface are in stiction, the surface contact patch and object move together as a rigid body. When the object is going to slip, some portion of the surface contact patch starts losing stiction with the object, and the motion of the contact surface is no longer a rigid body motion. 

\begin{figure}[ht]
    \centering
    \vspace{-.25 cm}
    \includegraphics[width=\columnwidth]{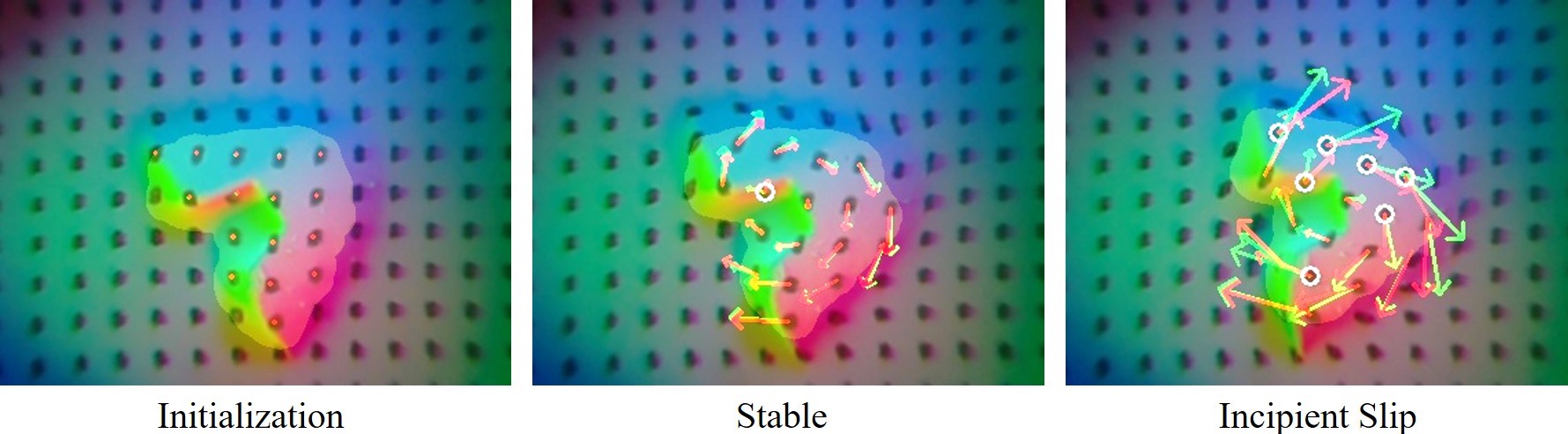}
    \vspace{-0.6 cm}
    \caption{The evolution from stable contact to incipient slip while the object is being rotated. We label the contact region with yellow color, ``real" marker tracking with a red arrow, ``estimate" marker tracking under rigid body motion constraints with green arrows, and slipping regions (markers) with a white circle.}
    \label{fig:slip}
    \vspace{-.25 cm}
\end{figure}

We first detect the contact region in the tactile image by selecting the region with a depth that exceeds a preset threshold. We track the markers' motion in the contact region as the ``real" motion of the sensor surface in contact. We then calculate the marker motion's rigid body transformation matrix in contact and get the ``estimated" motion of the sensor surface under rigid body constraints. A significant deviation between the ``real" and ``estimate" motion field represents incipient slip. Fig.~\ref{fig:slip} shows incipient slip detection with the tactile images captured while a chess piece is rotated in the grasp. 

\subsection{Dense Force Distribution}
\label{subsec:force}
To measure an accurate force field, we correspond the gel surface's motion field (measured by tracking the 2D marker motion) and the gel's deformation in the z-axis (depth image) to the force field paired with the FEM model of the gel. The method to do so, proposed by Ma et al. \cite{dense_force} and implemented in GelSlim 2.0, can be augmented using the depth reconstruction offered by the novel GelSlim 3.0 sensor. 

\begin{figure}[ht]
    \centering
    \vspace{-.3 cm}
    \includegraphics[width=\columnwidth]{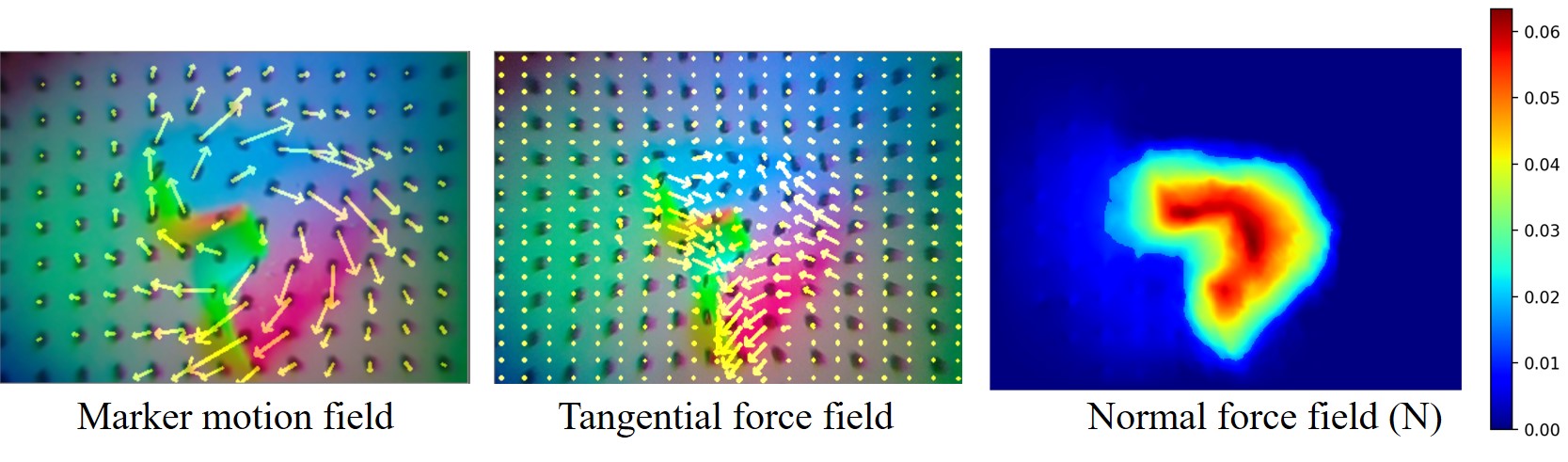}
    \vspace{-0.7 cm}
    \caption{The marker motion field, tangential force field and normal force field (N)}
    \label{fig:force}
    \vspace{-.5 cm}
\end{figure}

We discretize the gel using $m$ 8-node hexahedron elements, where the displacements of the 8 nodes represent the deformation of one FEM element. We calculate the displacement $\delta_x, \delta_y$ of the 4 nodes on the top by interpolating the marker motion field, the displacement $\delta_z$ with the depth map measurement. The $\delta_x, \delta_y$ need to be adjusted based on the viewing angle between the node-camera frame and $\delta_z$. We use standard FEM theory to calculate the stiffness matrix $K$  with the 8-node hexahedron elements and measured Young's Modulus and Poisson's ratio of the gel. The force field then can be directly calculated with the following equation:
\begin{equation}
  F = KU  
\end{equation}
where U is the displacement matrix of all the nodes which is directly observable. We show the marker motion field, dense shear force and the normal force field in Fig.~\ref{fig:force} when the sensor is holding a twisted chess piece. 

\section{Discussion}
\label{sec:discussion}
We compare the design components of our sensor to other prominent sensors in Table~\ref{table:comparison}. We find that our design provides a form factor with comparable thickness and price point to the smallest available sensor Digit while providing a sensing area that is twice as large. This is achieved while incorporating all the sensing modalities (high-resolution images, depth, and force) found in previous versions of GelSight and GelSlim. 

\subsection{Design Comparison, Limitations and Trade-offs}
\label{subsec:comparison}

\begin{table}[h]
\caption{Comparison of GelSlim 3.0, GelSlim 2.0, GelSight, Digit and Omnitatct. (*Considering the manufacturing of 1000 pieces)}
\label{table:comparison}
\resizebox{\columnwidth}{!}{%
\begin{tabular}{l|lllll}
                         & GelSlim 3.0 & GelSlim 2.0~\cite{dense_force} & GelSight~\cite{improved_gelsight}   & Digit~\cite{digit}     & Omnitact~\cite{omnitact}  \\
\hline
Size {[}mm{]}                   & 37X80x20    & 50x172x25   & 40x80x40  & 20x27x18 & 30x30x33 \\
Weight {[}g{]}                  & 45          & 222          & NA        & 20       & NA       \\
Sensing Field {[}mm$^{2}${]}    & 675         & 1200        & 252       & 304      & 3110       \\
Image Resolution                & 640x480     & 640x480     & 640x480 & 640x480  & 400x400  \\
Image FPS                       & 90          & 90          & 30        & 60       & 30       \\
Cost Components {[}\${]}        & 25*         & NA          & 30  & 15*      & 600     
\end{tabular}
}
\end{table}

Design, in general, is a compromise; the process and results of designing the novel GelSlim tactile finger are no exception. The tradeoffs demonstrated by the optimized sensor are generalizable to tactile sensors' design in other form factors. Thus they are important lessons to keep in mind for future designs. First and foremost is the use of planar sensing surfaces alongside photometric stereo. In this instance, the sensing surface was not used to shape the illumination in an effort to enable grasping from flush surfaces. Although we demonstrated the ability to integrate these two design features using shaping methods, it came at the cost of significantly constraining the design space. We would like to explore alternative methods of fulfilling the photometric stereo assumptions to achieve more compact integrations. For example, shaping the illumination directly at the source by using custom-designed LED microlenses.  

Next is the relationship between the camera module's size, the field of view, and the sensor's thickness. To further minimize the form factor, the field of view often must be reduced. We used a wide-angle fisheye lens as a workaround to this tradeoff, but this and other methods, for example utilizing more than one camera, often come at a processing cost. Most image processing software pipelines are sufficiently advanced such that the cost-reward relationship is favorable but may quickly diminish when using alternative workarounds, for example utilizing more than one camera to view the sensing surface. In this instance, the cost-reward relationship might be less favorable due to the amount of high-dimensional data that needs to be processed. 
    
Finally, and perhaps most important, is the influence of compactness on the second assumption of the photometric stereo algorithm. It is clear from our results that while interreflections can be minimized through shaping methods and absorptive coatings (in our case marker ink), they cannot be entirely eliminated, and thus they will add some residual noise within the reconstruction. While our design can capture texture information, the features have to be relatively large. This is in comparison with the previous versions of GelSight, which could easily reconstruct micro geometries and surface textures \cite{gelsight_texture,improved_gelsight,microgeometry,retrographic}. That being said, left unaccounted for, this interreflective behavior will inhibit the ability to further minimize the sensor thickness in future designs that aim to utilize photometric stereo depth reconstruction. 

\section{Open-Source Tactile-Sensing Finger}
\label{sec:source}
 On the GelSlim Github, we provide documentation, design schematics, and production files (.gbr, .step) for all of the requisite components needed to fabricate a sensor. We also provide a simple software GUI to interface with the sensor and a tutorial describing a data collection system's setup using a Raspberry Pi. The software provides easy access, and use of the three integrated tactile-sensing algorithms described (depth reconstruction, dense force distribution and incipient slip detection) and incorporates the ability to collect, measure and plot tactile imprint data in real-time. 


\section{Conclusions}
\label{sec:conclusion}
We present a novel GelSlim tactile finger design and develop the design using illumination design software and design for assembly techniques to analyze, explore, and optimize the finger within its design space. Specifically, we evaluate the effect of incorporating photometric stereo measurement techniques on the design space and demonstrate methods to optimize illumination shaping features to fulfill its assumptions. We develop a compact form factor with replaceable modules suited for grasping in cluttered bin picking scenarios that are easy to manufacture, use, and maintain.  We demonstrate that our sensor can utilize all of the measurement methods used by previous GelSight and GelSlim sensors, specifically 3D depth reconstruction, incipient slip and shear detection, and dense force distribution estimation. We compare our sensor to several other prominent tactile-sensing designs and discuss its limitations. We anticipate our future work to continue exploring and developing illumination shaping designs suited explicitly for analytic tactile measurement methods.

\section{Acknowledgment}
\label{sec:acknowledgment}
This work was supported by the NEC Corporation Fund for Research in Computers and Communications, and by the HKSAR Innovation and Technology Fund (ITF) ITS-104-19F. We thank Synopsys, Inc. for providing an educational license for the use of the LightTools Illumination Design Software. This article solely reflects the opinions and conclusions of its authors and not Synopsys or any other Synopsys entity. We thank Yu She for assisting with the fabrication of the elastomer.

\bibliographystyle{IEEEtran}
\bibliography{bibliography}

\appendix
 Simulation settings, optimization parameters and resultant configuration values can be found in the \textit{Simulation and Optimization} folder at \href{https://github.com/mcubelab/gelslim}{https://github.com/mcubelab/gelslim}.

\end{document}